%% file: AcceleratingSparsity.tex
\documentclass[runningheads]{llncs}
\usepackage{graphicx}
\usepackage{amsmath,amssymb} 
\usepackage{color}
\usepackage[width=122mm,left=12mm,paperwidth=146mm,height=193mm,top=12mm,paperheight=217mm]{geometry}

\usepackage{hyperref}       
\usepackage{url}            
\usepackage{booktabs}       
\usepackage{amsfonts}       
\usepackage{nicefrac}       
\usepackage{microtype}      
\usepackage{todonotes}
\usepackage{setspace}
\usepackage{framed}
\usepackage{multirow}
\usepackage{epstopdf}
\usepackage{cite} 
\usepackage{soul}           
\usepackage{enumitem}       

\usepackage{xcolor}

\newcommand{\SPARSITY}{2:4 }

\setuptodonotes{shadow, color=yellow!10!red!80}
\definecolor{nvidiagreen}{RGB}{116,183,27}
\definecolor{cardinal}{rgb}{0.77, 0.12, 0.23}
\definecolor{darkpastelred}{rgb}{0.76, 0.23, 0.13}
\definecolor{cornflowerblue}{rgb}{0.39, 0.58, 0.93}
\definecolor{cadmiumred}{rgb}{0.89, 0.0, 0.13}

\hypersetup{
     colorlinks   = true,
     citecolor    = cadmiumred, 
     linkcolor    = cadmiumred, 
}

\begin{document}

\pagestyle{headings}
\mainmatter
\title{Accelerating Sparse Deep Neural Networks}
\titlerunning{$\cdot$~Accelerating Sparse Deep Neural Networks~$\cdot$}
\authorrunning{$\cdot$~Accelerating Sparse Deep Neural Networks~$\cdot$}
\author{Asit Mishra, Jorge Albericio Latorre, Jeff Pool, Darko Stosic, \\ Dusan Stosic, Ganesh Venkatesh, Chong Yu, Paulius Micikevicius}
\institute{NVIDIA \\ \vspace{3mm} \texttt{\{asitm, jalbericiola, jpool, darkos, dstosic, chongy, pauliusm\}\\@nvidia.com}}

\maketitle

\small
\setlength{\abovecaptionskip}{-1pt}
\setlength{\belowcaptionskip}{-8pt}

\begin{abstract}

\input{abstract}

\end{abstract}

\section{Introduction} 
\label{sec:into}

\input{intro}

\section{Related Work} 
\label{sec:related}
\input{prior_work}

\section{Sparsity Support in the NVIDIA Ampere Architecture} 

\label{sec:hw}
\input{hardware}

\section{Network Pruning Workflow} 
\label{sec:sw}
\input{software}

\section{Results} 
\label{sec:res}
\input{results}

\section{Conclusions and Future Work} 
\label{sec:conclusion}
\input{conclusion}

\clearpage
\bibliographystyle{unsrt}
\bibliography{AcceleratingSparsity}  

\end{document}

%% file: abstract.tex
As neural network model sizes have dramatically increased, so has the interest in various techniques to reduce their parameter counts and accelerate their execution. An active area of research in this field is sparsity -- encouraging zero values in parameters that can then be discarded from storage or computations. While most research focuses on high levels of sparsity, there are challenges in universally maintaining model accuracy as well as achieving significant speedups over modern matrix-math hardware. To make sparsity adoption practical, the NVIDIA Ampere GPU architecture introduces sparsity support in its matrix-math units, Tensor Cores. We present the design and behavior of Sparse Tensor Cores, which exploit a \SPARSITY (50\%) sparsity pattern that leads to twice the math throughput of dense matrix units. We also describe a simple workflow for training networks that both satisfy \SPARSITY sparsity 
pattern requirements \textit{and} maintain accuracy, verifying it on a wide range of common tasks and model architectures. This workflow makes it easy to prepare accurate models for efficient deployment on Sparse Tensor Cores.

%% file: intro.tex
In the area of Deep Learning, using larger neural network models typically leads to higher accuracy for various tasks~\cite{hestness2017scaling,hestness2019humanlevel,patwary2018scale,meshtf2018}.
Modern state-of-the-art models can consist of hundreds of billions of parameters and require trillions of compute operations per input sample.
Pruning of neural network parameters has emerged as an important technique to reduce model sizes and compute requirements at inference time. 

Pruning accomplishes this by pushing certain parameter values to zero, inducing sparsity in a model~\cite{strom97sparseconnection,Han2016DSDRD,han-pruning,mao2017exploring,zhu2017prune,he2017channel,molchanov2016pruning,liu2018rethinking,LTH}.
However, existing pruning methods can struggle to simultaneously maintain model accuracy and gain inference performance (speed).
Fine-grained sparsity maintains accuracy but poorly utilizes memory accesses and fails to take advantage of modern vector and matrix math pipelines, thus it does not outperform traditional dense models on processor architectures such as GPUs. 
Coarse-grained sparsity can better utilize processor resources but fails to maintain accuracy beyond moderate sparsity ratios.

In this paper, we describe a \SPARSITY (read as ``two-to-four") sparsity pattern that halves a model's parameter count, requiring that every group of consecutive four values contains at least two zeros.
We also describe a workflow to prune traditional, dense models for this pattern while maintaining their accuracy.
This workflow prunes weights of a densely-trained model once, then repeats the training session with a fixed sparsity pattern using the same hyper-parameters as in the original training session.
Furthermore, we describe Sparse Tensor Cores, introduced in the NVIDIA Ampere GPU architecture~\cite{ampere}, to accelerate operations on \SPARSITY sparse matrices.
Sparse Tensor Cores double math throughput for matrix-multiply operations when the first argument is a compressed 2:4 sparse matrix.
Matrix multiplication is a compute primitive behind math-intensive neural network operations such as convolutions, linear layers, recurrent cells, and transformer blocks.

The contributions of this paper include: (1) a fine-grained 2:4 structured pruning approach, (2) a compression format to efficiently store such pruned tensors in memory, (3) hardware architecture to accelerate sparse matrix multiplication involving a 2:4 sparse tensor, and (4) an empirically-verified workflow that re-trains pruned weights to eliminate accuracy loss in many standard networks.
We summarize prior sparsity research in Section~\ref{sec:related}.
The 2:4 pattern, compressed storage format required by Sparse Tensor Cores, and Sparse Tensor Core operation are detailed in Section~\ref{sec:hw}.
Section~\ref{sec:sw} describes the methodology for training neural networks to have 2:4 weight sparsity so that inference can be accelerated.
In Section~\ref{sec:res}, we provide an empirical study on a variety of popular tasks and neural network architectures highlighting the universality of our proposed workflow in maintaining model accuracy while not having to change any hyper-parameters.
Section~\ref{sec:conclusion} concludes the paper with a summary and directions for future work. 

%% file: prior_work.tex
Neural network model pruning approaches can be grouped into the following categories:
\begin{itemize}
\item train and prune a dense model, then fine-tune the remaining weights in the model to recover accuracy,
\item train a dense model with gradual pruning to obtain a sparse model,
\item train a sparse model with a sparsity pattern selected \textit{a priori}, or
\item train a sparse model with a sparsity pattern determined based on trained dense version.
\end{itemize}

Fine-tuning methods prune fully-trained dense weights and continue to fine-tune
the remaining weights for additional training samples.
Different approaches in this space can be distinguished by the pruning method, pruning schedule, sparsity structure, and fine-tuning schedule used during the fine-tuning phase.  
Pruning methods typically eliminate weights using weight magnitude based metrics~\cite{han-pruning}
or some salience-based criteria~\cite{Molchanov_2019_CVPR}).
A variety of pruning schedules have been proposed -- single step~\cite{lee2018snip} and
gradual/iterative~\cite{he2017channel,molchanov2016pruning,yao2018balanced}. Iterative methods prune weights gradually over a number of steps, in each step eliminating either a fixed number of weights~\cite{han-pruning,Han2016DSDRD} or choosing a fraction of weights based on an analytical function~\cite{zhu2017prune}. Furthermore, the pattern used to prune models may adhere to a specific structure~\cite{yao2018balanced,wang2018structured,anwar2015structured,he2017channel} 
or follow no structure at all~\cite{han-pruning,molchanov2016pruning}. Structured pruning removes parameters in groups (entire filters, channels, etc.) in order to exploit hardware and software optimized for dense computation. However, at higher levels of sparsity these pruning methods lose model accuracy~\cite{he2017channel,lin2020filter,gamboa2020campfire,DBLP:conf/micro/ZhuZG019,DBLP:journals/corr/WenWWCL16}. For example, ResNet-50 can see a 2$\times$ speedup through channel pruning, but close to 1.5\% 
accuracy is lost~\cite{he2017channel}. Unstructured pruning eliminates individual parameters without any regard to the resulting pattern. Networks pruned with unstructured sparsity tend to retain more accuracy than similarly sized networks pruned with structured sparsity, but they rarely fully utilize the underlying hardware capabilities. For example, model parameters can be pruned by 
nearly 13$\times$ with no loss in accuracy, but the pruning pattern is not 
conducive to hardware acceleration~\cite{han-pruning}. Hence, the performance
benefit with such unstructured pruning approaches is negligible and at times 
negative, even when pruning rate is 
high (e.g. 95\%)~\cite{DBLP:journals/corr/WenWWCL16}. 
The difficulty of getting inference speed benefits with unstructured patterns is even more pronounced on processor architectures with matrix pipelines, such as GPUs and TPUs (Tensor Cores and systolic arrays, 
respectively). No universal fine-tuning schedule has been proposed yet -- schedules often vary from model to model (for example,using 10 epochs of fine-tuning for one network but requiring 20 epochs for another network model\cite{he2017channel}).

Some approaches train models with a fixed sparsity pattern starting from the random initialization of the network, but the pattern is computed based on a fully trained~\cite{LTH} or partially trained~\cite{frankle2019stabilizing,renda2020comparing} dense model of the same architecture. These approaches maintain accuracy and achieve 80\%-99\% sparsity. While run-time is not discussed in these works, given the use of unstructured sparsity, it is unlikely for these pruned models to efficiently utilize modern matrix processors.

Sparse models can also be obtained by starting from randomly-initialized dense models and gradually introducing sparsity as training progresses~\cite{zhu2017prune,narang2017exploring,gamboa2020campfire,narang2017blocksparse,elsen2019fast,park2018squantizer,DBLP:journals/corr/abs-1802-08435,DBLP:journals/corr/abs-1911-11134}. Typically, 70\%-99\% unstructured sparsity is targeted to prepare models for inference. As described above, these models struggle to outperform dense models due to their unstructured sparsity. Furthermore, to maintain accuracy, model and training hyper-parameters are often modified, and these modifications can be quite model-specific. For example, in~\cite{narang2017exploring},
the hidden size of a recurrent network was increased by 1.75$\times$ while pruning to 95\% sparsity. In another case~\cite{DBLP:journals/corr/abs-1911-11134}, the training schedule was extended for some networks by 5$\times$ in order to reach the same accuracy as the dense model.

Finally, there have been proposals to design sparse model architectures, aimed at efficient hardware utilization during training and inference. Most prominent example of this category is block sparsity, where sparsity is introduced at a block level allowing good utilization of matrix math pipelines~\cite{Gray2017GPUKF}. However, similar to approaches discussed above, in order to maintain accuracy, one has to increase the hidden size compared to the dense model. Block sparsity has found use for cases where using a larger hidden size enables higher accuracy but is impractical with dense models.
Additionally, there has been work investigating fine-grained structured sparsity and motivating the need to prune in a fine-grained pattern that is conducive to hardware acceleration~\cite{yao2018balanced,AccAwarePruning}.
Key points are optimized GPU kernels to speed up such pruned models on CUDA cores~\cite{yao2018balanced} and the benefits of custom hardware to speed up structurally pruned models~\cite{AccAwarePruning}.

\subsubsection{Summary:} While a variety of approaches for pruning neural networks to high degrees of sparsity have been proposed, no one method has been described to maintain accuracy while achieving inference speedup. 
Fine-tuning workflows provide inconclusive results on which pruning schemes to use (e.g. magnitude or heuristic based weight pruning, prune partially trained or fully trained weights), what pruning method to follow (e.g. one-shot or iterative pruning, layer by layer pruning or whole network pruning.) and what fine-tuning schedule to use (e.g. how many epochs of fine-tuning, what learning rate to use, etc.). In short, extracting performance from hardware by pruning networks while maintaining accuracy and using a fine-tuning workflow that is consistent across a variety of networks is still an open problem.
This motivated our search for a sparsity pattern that enables hardware acceleration as well as maintains accuracy with a workflow applicable over a wide range of neural network tasks and models.

%% file: hardware.tex
We introduce \SPARSITY sparsity to address the challenges of adopting sparsity outlined in the last section. The \SPARSITY pattern mandates that for each group of 4 values, at least 2 must be zero. This leads to 50\% sparsity, which makes maintaining accuracy without hyper-parameter exploration much more practical than, say, 80\% sparsity. When applied to a matrix, the \SPARSITY pattern has the following benefits over alternative sparsity approaches:
\begin{itemize}[leftmargin=0.45in]
    \item efficient memory accesses
    \item a low-overhead compressed format
    \item 2x math throughput increase on the NVIDIA Ampere GPU architecture
\end{itemize}The pattern and compressed format are detailed in Section~\ref{sparsity_pattern}, while Sparse Tensor Cores that take advantage of this format are described in Section~\ref{sparse_tensorcore}.

\begin{figure*}[thbp]
    \centering
    \includegraphics[width=0.75\textwidth]{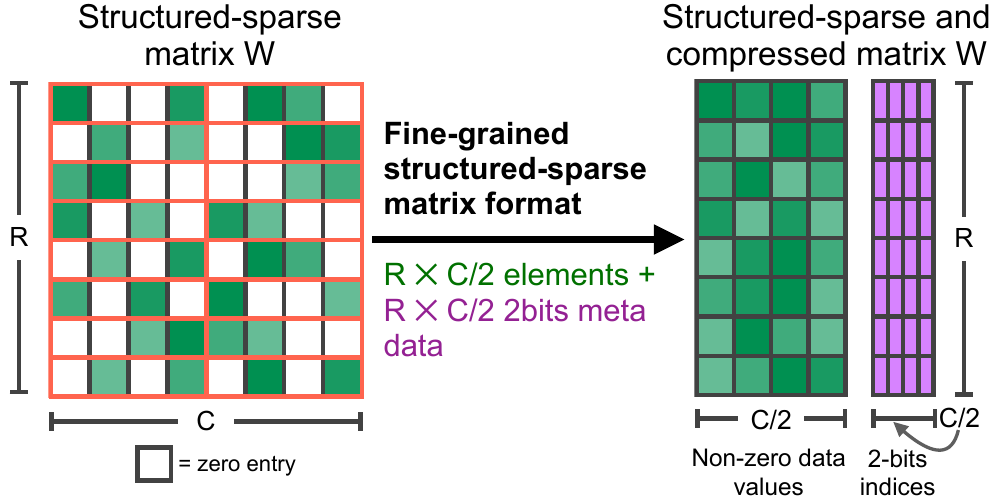}
    \caption{Structured-sparse matrix (W) storage format. The uncompressed matrix is of dimension R $\times$ C and the compressed matrix is of dimension R $\times$ $\frac{C}{2}$ .} 
     \label{fig:storage_format}
\end{figure*}

\subsection{\SPARSITY Sparsity and Its Benefits} \label{sparsity_pattern}

An example of a matrix that satisfies \SPARSITY sparsity pattern requirement is shown in Figure~\ref{fig:storage_format}.
With this pattern, only the 2 nonzero values in each group of 4 values need to be stored.
Metadata to decode compressed format is stored separately, using 2-bits to encode the position of each nonzero value within the group of 4 values. 
For example, metadata for the first row of matrix in Figure~\ref{fig:storage_format} is $[[0, 3], [1, 2]]$.
Metadata information is needed to fetch corresponding values from the second matrix when performing matrix multiplication.
Note that for a group of 4 values having more than 2 zeros, the compressed format will still store 2 values to maintain a consistent format.

\subsubsection{Efficient memory accesses:} Unstructured sparsity patterns lead to poor utilization of cache lines when accessing memory, thus under utilizing memory bandwidth. Furthermore, unstructured patterns commonly use CSR/CSC/COO storage formats~\cite{storageformat}, which lead to data-dependent accesses, thereby increasing latency for matrix reads.
In contrast, \SPARSITY sparsity has the same level of sparsity at every sub-block of the larger matrix, which enables hardware to fully-utilize large memory reads.
Similarly, since the sparsity is constant across the matrix, there is no indirection required; a nonzero value's position in memory can be determined from the compression rate directly.

\subsubsection{Compressed format efficiency:} Using CSR format for unstructured sparsity can introduce storage overhead due to metadata of up to 200\% (consider 8b quantized weight values for inference: column-index for the value would require 16-bits or more for even modestly-sized matrices). Due to its 4-value block size, the \SPARSITY sparse storage format (shown in Figure~\ref{fig:storage_format}) requires only 2-bits metadata per value, limiting storage overhead to 12.5\% and 25\% for 16b and 8b values, respectively. For 16-bit operands, storing a sparse tensor in compressed format leads to \texttt{$\sim$}44\% savings in storage capacity:
4 dense elements require 4$\ast$16 = 64-bits of storage while \SPARSITY sparsity leads to 2$\ast$16-bits + 2$\ast$2-bits = 36-bits to store the two non-zero elements. For 8-bit operands, storing in compressed format saves \texttt{$\sim$}38\% in memory capacity and bandwidth compared to the dense tensor. 

\subsection{Structured-Sparse GEMM on Tensor Cores} \label{sparse_tensorcore}

Tensor Cores, first introduced in the NVIDIA Volta GPU architecture, accelerate matrix-multiply-and-accumulate (MMA) instructions that are fundamental to neural network layers involving math operations such as convolutions, linear layers, recurrent cells, and transformer blocks.

\begin{figure*}[tb]
    \centering
    \includegraphics[width=0.975\textwidth]{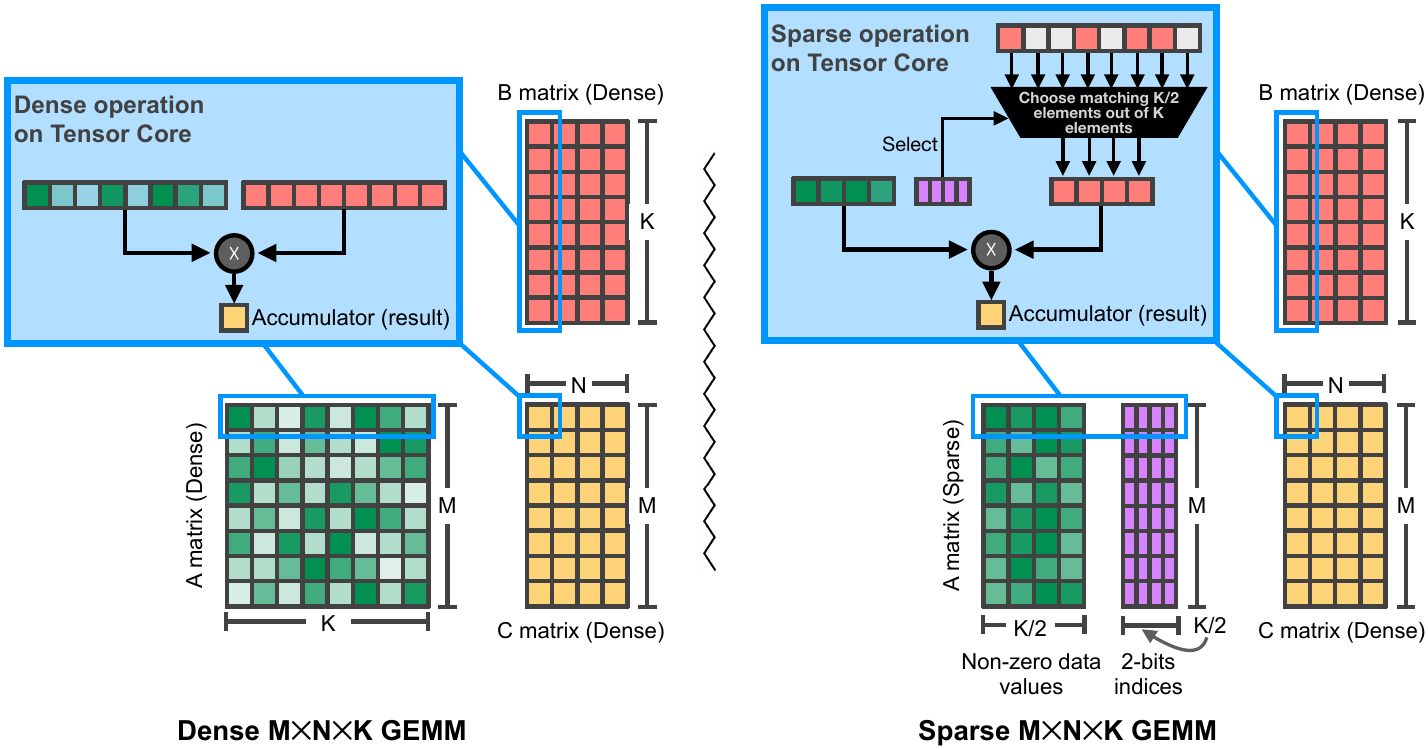}
    \caption{
    Mapping a M$\times$N$\times$K GEMM onto a Tensor Core.
    Dense matrix $A$, of size M$\times$K, (left side) becomes M$\times\frac{K}{2}$ (right side) after pruning with \SPARSITY sparsity.
    Sparse Tensor Core hardware selects only the elements from B that correspond to the nonzero values in A, skipping the unnecessary multiplications by zero.
    In both dense and sparse GEMMs, B and C are dense K$\times$N and M$\times$N matrices, respectively.}
    \label{fig:dp_engine_sparse_mat}
\end{figure*}

\begin{table}[tp]
\setlength{\tabcolsep}{10pt}
\centering
\caption{A100 Tensor Core input/output formats (numeric precision)
and performance in tera-operations per second (TOPS).}
\label{tab:tops_perf}
\renewcommand{\arraystretch}{1.10}
\begin{tabular}{@{}cccc@{}}
\textbf{Input Operands} & \textbf{Accumulator} & \textbf{Dense TOPS} & \textbf{Sparse TOPS} \\
\midrule
FP32    &   FP32    &   19.5    &   -     \\
TF32    &   FP32    &   156     &   312   \\
FP16    &   FP32    &   312     &   624   \\
BF16    &   FP32    &   312     &   624   \\
FP16    &   FP16    &   312     &   624   \\
INT8    &   INT32   &   624     &   1248  \\
\bottomrule
\vspace{-10mm}
\end{tabular}
\end{table}

The NVIDIA Ampere GPU architecture extends the Tensor Cores to also handle \SPARSITY sparsity by allowing the first argument be stored in the sparse format described in Section~\ref{sparsity_pattern}.
Thus, Sparse Tensor Cores perform sparse matrix $\times$ dense matrix = dense matrix operation (the second input matrix and the output matrix are dense).
Figure~\ref{fig:dp_engine_sparse_mat} shows how a \SPARSITY sparse GEMM operation is mapped to Tensor Cores. 
50\% sparsity on one of the operands halves the required multiply-and-add operations, resulting  in (up to) a 2$\times$ performance increase over equivalent dense GEMMs. Sparse Tensor Cores support FP16, BF16, and 8b-integer input/output types. Furthermore, TF32 mode is supported for FP32 input/output but the pattern becomes 1:2 sparse. Peak dense and sparse Tensor Core throughputs are shown in Table~\ref{tab:tops_perf}.

It is the application's responsibility to ensure that the first operand is a matrix stored in the compressed \SPARSITY format. cuSPARSELt and other libraries provide  APIs for compression and sparse math operations, while, starting in version 8.0, the TensorRT SDK performs these functions for 2:4 sparse weights automatically. NVIDIA libraries require that input dimensions of a sparse matrix multiplication be multiples of 16 and 32 for 16-bit (FP16/BF16) and 8b-integer formats, respectively.

\begin{figure*}[tb]
    \centering
    \includegraphics[width=0.75\textwidth]{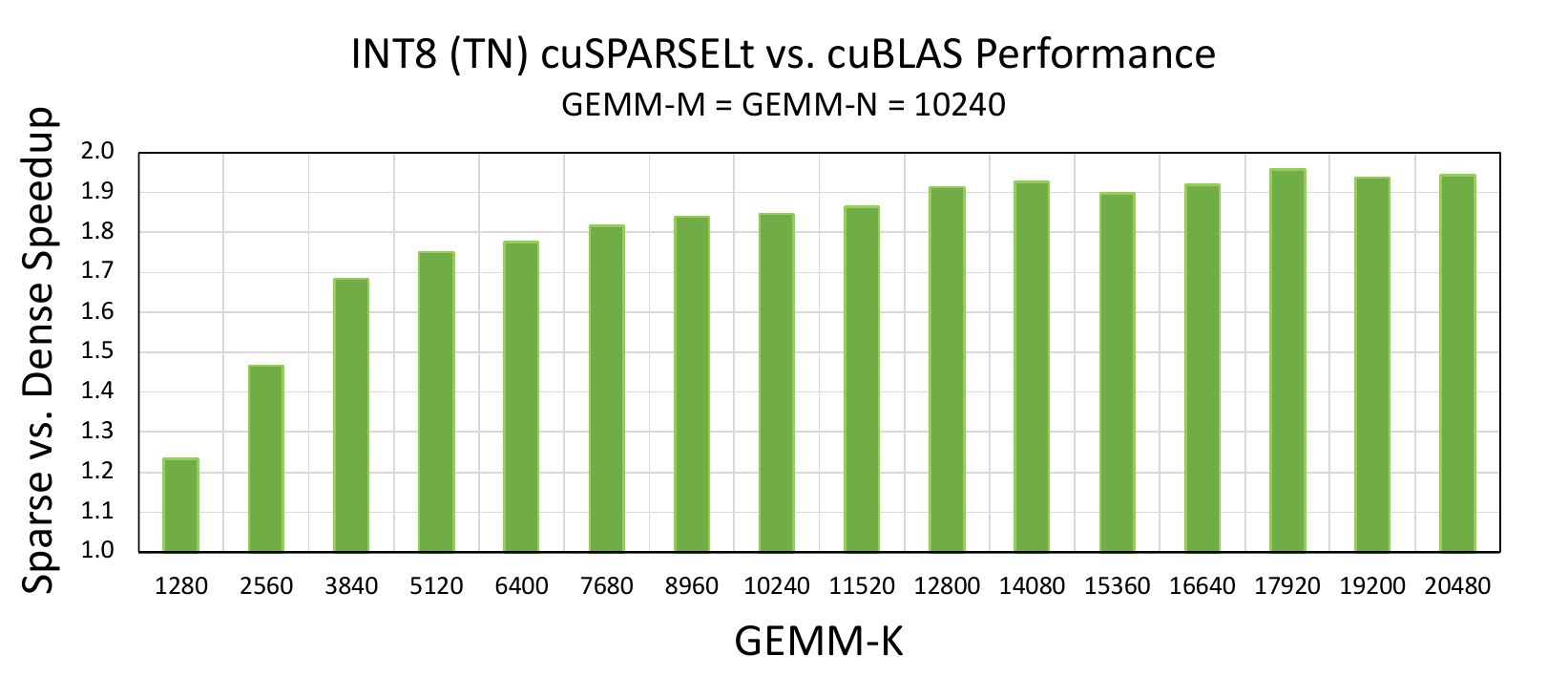}
    \caption{Comparison of sparse and dense INT8 GEMMs on NVIDIA A100 Tensor Cores. Larger GEMMs
    achieve nearly a 2$\times$ speedup with Sparse Tensor Cores.}
    \label{fig:gemmperf}
\end{figure*}

Speedups that \SPARSITY sparse matrix multiplications achieve over dense multiplications depend on several factors, such as arithmetic intensity and GEMM dimensions.
Figure~\ref{fig:gemmperf} shows speedups achieved over a sampling of GEMM dimensions (cuSPARSELt\footnote{https://docs.nvidia.com/cuda/cusparselt/index.html} cuBLAS\footnote{https://docs.nvidia.com/cuda/cublas/index.html} libraries were used for the sparse and dense GEMMs, respectively).
As larger GEMMs tend to have higher arithmetic intensity, they get closer to the $2\times$ speedup afforded by Sparse Tensor Cores.
For language modeling networks, N is often the sequence length times the batch size: for a sequence length of 256, one would need a batch size of 40 to see this plot with N equal to 10K.
M and K are related to the hidden dimensions of the layers in the network, which is typically scaled up to increase network accuracy; GPT-3~\cite{GPT-3}, for example, uses a hidden size of 12,288.

%% file: software.tex
In this section, we describe a workflow that prunes a network with the \SPARSITY sparsity pattern, maintains original accuracy, and avoids any hyper-parameter searches. Since our aim is to reduce neural network size and run-time at deployment, we trade a higher training cost for a simple and general workflow -- the additional training cost can be amortized over the deployment lifetime of days to months.

\begin{figure*}[th]
    \centering
    \includegraphics[width=0.70\textwidth]{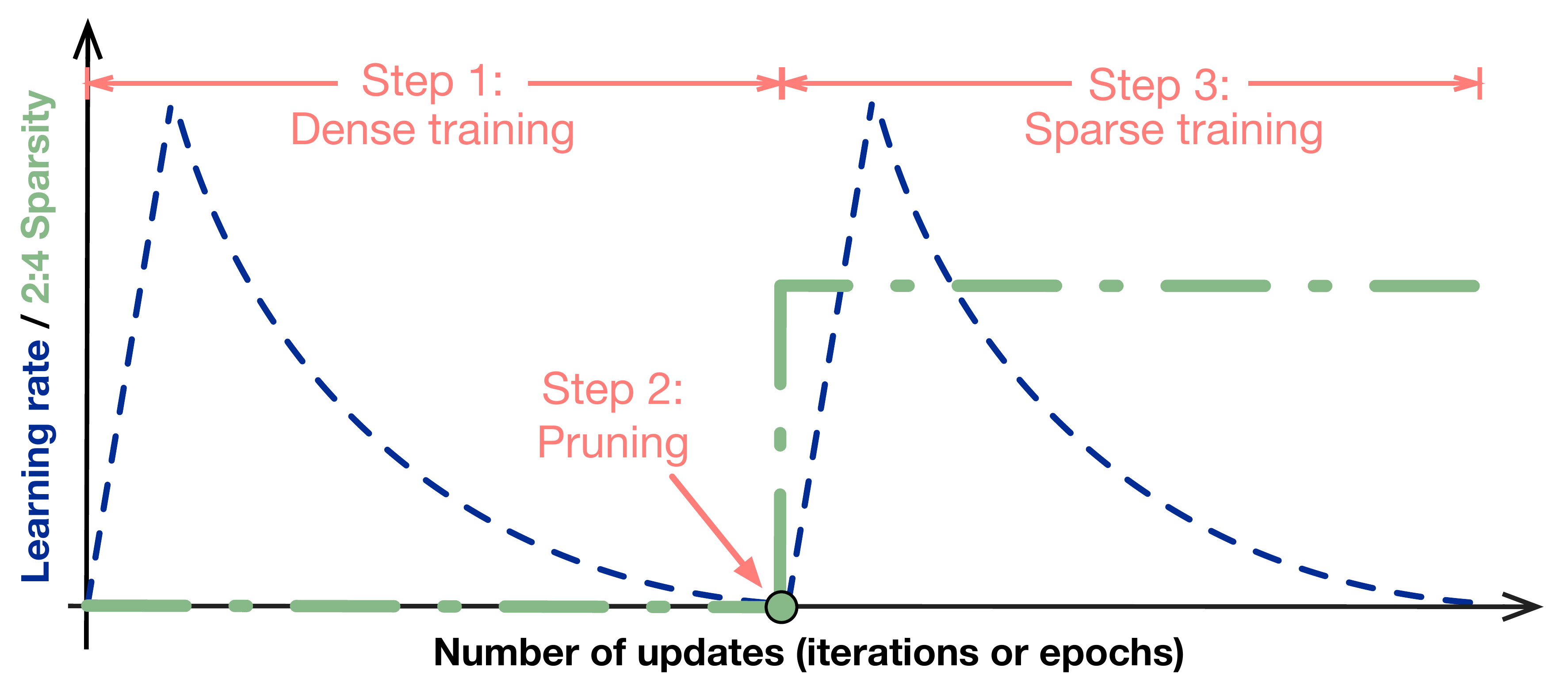}
    \caption{A typical network training process adapted for \SPARSITY sparsity: dense training (step 1) is followed by a pruning stage (step 2) and a second, identical training stage (step 3).}
    \label{fig:recipesimple}
\end{figure*}

\subsection{The Basic Workflow} \label{recipe_section}

While our proposed workflow trains a network twice, it achieves universality - as we will show, it can be applied across a wide range of neural network architectures and tasks. It follows the basic train, prune, and fine-tune approach:

\begin{enumerate}[leftmargin=0.45in]
    \item Train a model without sparsity,
    \item Prune the model in a \SPARSITY sparse pattern,
    \item Retrain the pruned model:
    \begin{itemize}
        \item initialize the weights to the values from Step 2,
        \item use the same optimizer and schedule (learning-rate, schedule, number of epochs, etc.) as in Step 1,
        \item maintain the sparsity pattern computed in Step 2.
    \end{itemize}
\end{enumerate}

This workflow is implemented in the Automatic SParsity (ASP)~\cite{ASP} library for PyTorch and is illustrated in Figure~\ref{fig:recipesimple}, where the two training stages are identical in learning rate schedule and length and are separated by the one-shot pruning step.
While Step 1 is straightforward, some details about Steps 2 and 3 follow.

\subsubsection{Step 2: Weight pruning:} 
\begin{figure*}[thb]
    \centering
    \includegraphics[width=0.75\textwidth]{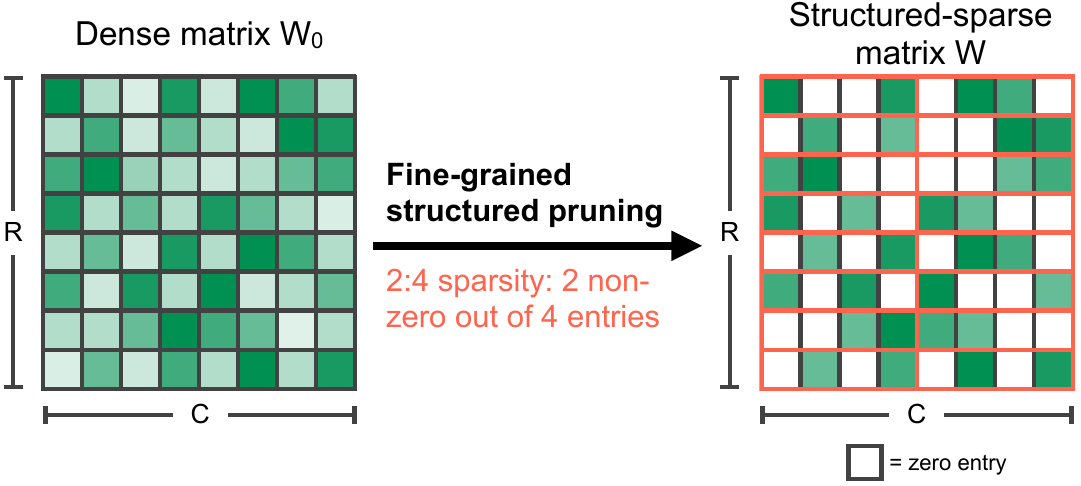} 
    \caption{Step 2 of the workflow modifies the weights in the dense matrix (W$_0$) on the left to conform to the \SPARSITY constraint, yielding the sparse matrix (W) on the right.}
     \label{fig:prune}
\end{figure*}

At its simplest, the pruning step removes the two smallest weights in each group of four to meet the \SPARSITY pattern requirement, as illustrated in Figure~\ref{fig:prune}. We found using the magnitude criteria sufficient, but one could consider other metrics, such as output activation similarity or instantaneous gradients.

\vspace{-6mm}

\vspace{-10mm}
\begin{figure*}[th]
    \centering
    \includegraphics[width=0.9\textwidth]{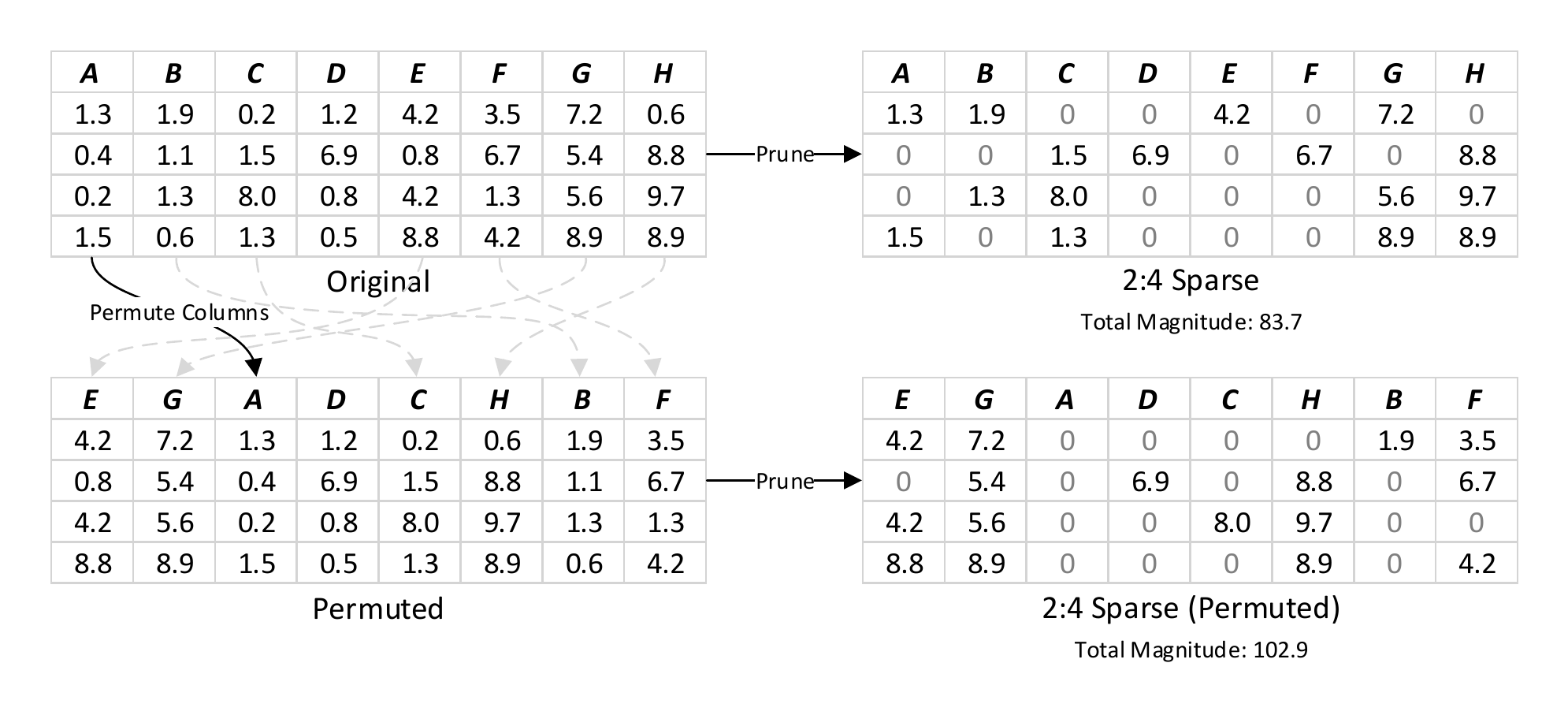} 
    \vspace{-1mm}
    \caption{Permuting columns of a weight matrix prior to pruning the matrix can reduce the effect of the \SPARSITY sparsity constraint on weight magnitude.}
     \label{fig:permute}
\end{figure*}

At this point in the workflow, the opportunity exists to change the layout of the network's weights using channel permutations to minimize the impact of the pruning step.
Figure~\ref{fig:permute} shows how this works on a random dense matrix (top-left).
When this matrix is pruned with the \SPARSITY sparse constraint (top-right), some relatively large values are lost, resulting in a final total weight magnitude of 83.7.
By first permuting columns of this weight matrix to distribute the large values more evenly (lower-left), they are preserved after pruning (lower-right), for a final total weight magnitude of 102.9.

Since permutations are applied to columns of weights (the $A$ matrix in Figure~\ref{fig:dp_engine_sparse_mat}), the rows of activations ($B$) must be similarly permuted to maintain the same result for the GEMM operation.
We accomplish this by permuting the \textit{rows} of the weights used to produce those activations, typically the weights of the previous layer.
For convolutions, permutations are applied to the input channel dimension, which becomes a component of the weight matrix's column dimension.
This process does not change the computations performed or the network's results and incurs no runtime overhead.

For the majority of networks tested, these permutations are not required - step 2 is as simple as enforcing the 2:4 constraint on the weight tensors as they are.
However, some networks are designed with efficiency in mind and begin with very few parameters; simply pruning and fine-tuning these models may still result in accuracy loss.
Channel permutations make the most of each nonzero parameter and allow \SPARSITY sparsity to maintain accuracy for these parameter-efficient networks, as shown in Table~\ref{tab:permutations}. Details about finding quality permutations will be presented separately.

\subsubsection{Step 3: Sparse retraining:}
We use retraining to recover model accuracy lost when half of the weights are removed by Step 2 of the workflow.
To avoid any hyper-parameter search, we simply repeat the training session from Step 1, starting with weights from Step 1 rather than a random initialization.
It is important to reset all hyper-parameters and optimizer state, such as momenta, etc.
Any weight removed in Step 2 should retain its zero value in order to maintain the \SPARSITY sparsity pattern.

\subsection{Layers to Prune} 
In our studies, we prune only layers that have learnable parameters and lead to a GEMM-like operation during the neural network's execution on hardware.
Such layers include convolution, fully-connected, and recurrent layers.
We do not prune layers with inner dimensions (GEMM-K for fully-connected or recurrent layers and C$\times$R$\times$S for convolution layers) that are not multiples of 16 and 32 for 16-bit floating point and 8b-integer formats, respectively.
We also do not prune embedding layers (typical in language processing tasks) since these layers effectively implement a lookup table.
Further, since our goal is to speed up inference, we do not prune layers that are involved only in the training phase and not in the inference phase of the network.
Such layers include language-modeling heads used during training in language processing networks (like BERT~\cite{bert}) which are then replaced by task-specific heads during inference, auxiliary classifiers used in Inception networks~\cite{DBLP:journals/corr/SzegedyLJSRAEVR14} which are removed altogether for deployment, and the entirety of discriminator networks used in adversarial training of generative networks (GANs).

\subsection{Applying the Workflow to Models Trained in Multiple Phases}
For models that are trained in a single phase, the application of the workflow from Section~\ref{recipe_section} is straightforward.
Examples of single-phase training include image classification networks trained on the ILSVRC2012~\cite{ILSVRC15} dataset, language translation networks trained on a single dataset, etc.
However, when networks are trained in multiple phases, we can consider how many and which phases to consider for Steps 1 and 3 of the workflow.
For example, object detection networks are often trained in 2 phases: first the backbone is trained on ILSVRC2012, then the detector heads are added and the model is trained for detection on COCO~\cite{coco} dataset.
Another example is question answering networks, such as BERT, that are first trained for language modeling and \textit{then} trained for question answering on another dataset.
We break such scenarios down into two categories:

\subsubsection{Cases that require pruning and retraining the first phase:}
\begin{figure*}[tb]
    \centering
    \includegraphics[width=1.0\textwidth]{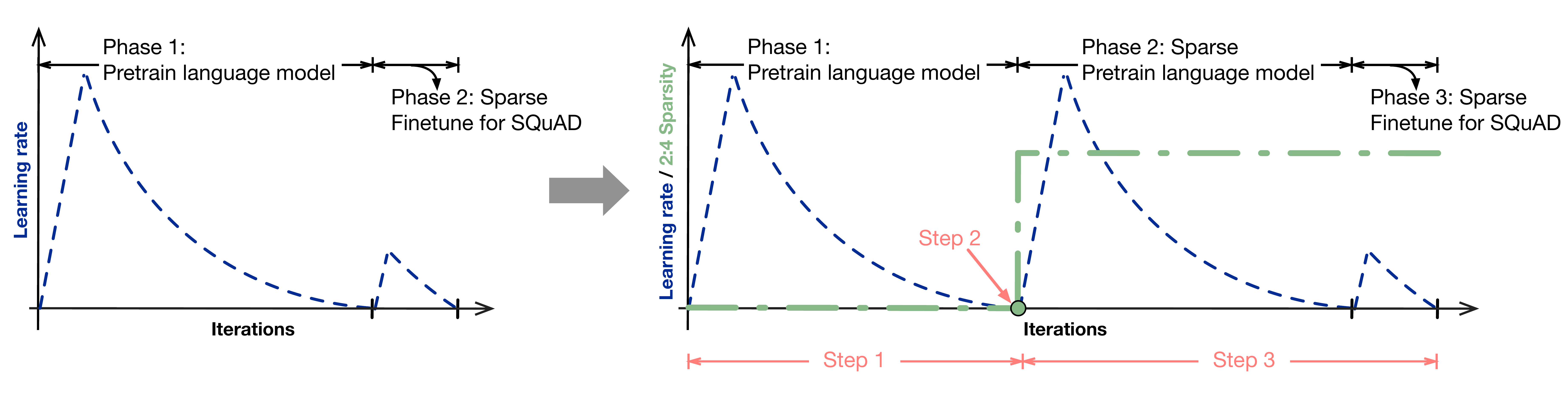}
    \caption{BERT's downstream tasks are too small recover accuracy lost by pruning the language model, so the pre-training is repeated with sparsity before fine-tuning for the downstream task.}
    \label{fig:recipebert}
\end{figure*}

In some cases where the second phase trains on a small data set, we have observed that the pruned network does not go through enough updates to recover accuracy after pruning if only the second phase is used for the retraining step of the workflow.
An example of this case is a language model like BERT, which is pre-trained on a very large dataset and then fine-tuned on a much smaller dataset for downstream tasks.
The solution to this problem is to simply prune and retrain after the \textit{pre-training} step, as shown in Figure~\ref{fig:recipebert}.
Then, the fine-tuning for the downstream task starts with a sparse and retrained model and simply maintains the sparsity pattern. 

\begin{figure*}[tb]
    \centering
    \includegraphics[width=0.65\textwidth]{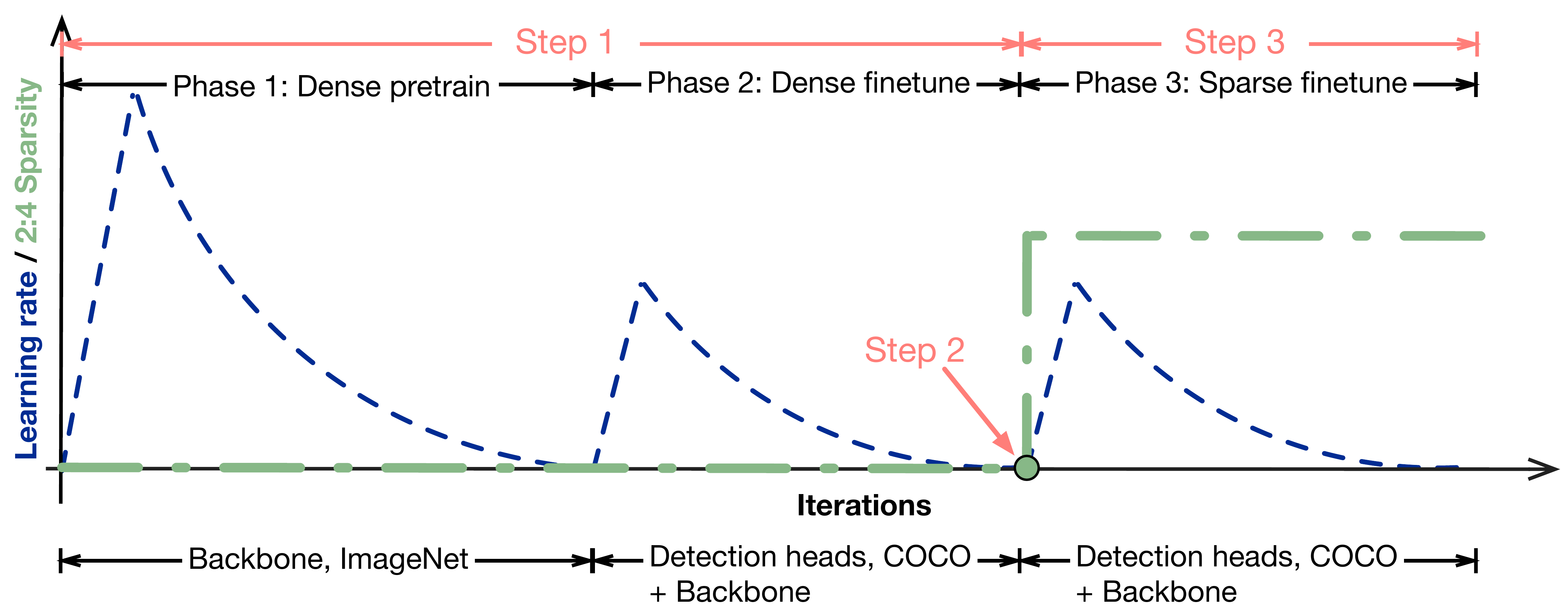}
    \caption{A detection network's fine-tuning of the detection heads is enough to recover accuracy lost by pruning the backbone, so the pruning step is inserted after the full network is trained, and only the fine-tuning is repeated.}
    \label{fig:recipedetect}
\end{figure*}

\subsubsection{Cases that can prune and retrain only the second phase:}
In contrast to language models, common object detection tasks fine-tune with a large-enough data set to \textit{not} require repeating the backbone's pre-training. In this case, it is sufficient to train a dense backbone and fine-tune detection or segmentation heads. At this point, after all the weights have been trained, they can be pruned and the fine-tuning repeated, as shown in Figure~\ref{fig:recipedetect}. It is important to note that the task-specific heads have to be fine-tuned before they can be pruned.

These examples show different types of two-phase training; the same principles can be applied to training sessions consisting of more than two phases.

\subsection{Combining Sparsity and Quantization}
Quantization is a popular technique to accelerate neural network inference -- by adapting the network to use narrower integer types, for example INT8, we can both reduce memory bandwidth pressure and benefit from higher throughput math pipelines~\cite{haoly}.
Quantizing a network typically starts with a network trained in floating point, then calibration is applied to determine the parameters for replacing floating point values and math with low-bit integer ones.
While some networks retain accuracy immediately after quantization, others require fine-tuning to recover lost accuracy. For both types of networks our recommendation is to apply quantization calibration (and potentially fine-tuning) \textit{after} a network has been pruned (and retrained) for sparsity.

%% file: results.tex
We evaluate the workflow proposed in Section~\ref{sec:sw} across a range of problem domains, tasks, and neural network architectures. For training each of the networks, we use hyper-parameters and training details mentioned in the papers introducing the network architecture and/or popular public repositories of network implementations.
We examine model accuracy for both floating point networks as well as their quantization to INT8.

\vspace{-3mm}
\begin{table}[htbp]
\setlength{\tabcolsep}{10pt}
\centering
\caption{Top-1 accuracy of image classification networks evaluated on the ImageNet ILSVRC2012 dataset 
with \SPARSITY sparsity.}
\label{tab:classification_nets}
\renewcommand{\arraystretch}{1.1}
\begin{tabular}{@{}llll@{}}
\multirow{3}{*}{\textbf{Network}} & \multicolumn{3}{c}{\textbf{Accuracy}} \\ \cmidrule(l){2-4}
                         & \textbf{Dense}   & \textbf{Sparse}   & \textbf{Sparse} \vspace{-1.0mm} \\
                         & \textbf{FP16}    & \textbf{FP16}     & \textbf{INT8}    \\
\midrule
ResNet-34                     & 73.7  & 73.9        & 73.7        \\
ResNet-50                     & 76.1  & 76.2        & 76.2        \\
ResNet-50 (SWSL)              & 81.1  & 80.9        & 80.9        \\
ResNet-101                    & 77.7  & 78.0        & 77.9        \\
ResNeXt-50-32x4               & 77.6  & 77.7        & 77.7        \\
ResNeXt-101-32x16             & 79.7  & 79.9        & 79.9        \\
ResNeXt-101-32x16 (WSL)       & 84.2  & 84.0        & 84.2        \\
DenseNet-121                  & 75.5  & 75.3        & 75.3        \\
DenseNet-161                  & 78.8  & 78.8        & 78.9        \\
Wide ResNet-50                & 78.5  & 78.6        & 78.5        \\
Wide ResNet-101               & 78.9  & 79.2        & 79.1        \\
Inception v3                  & 77.1  & 77.1        & 77.1        \\
Xception                      & 79.2  & 79.2        & 79.2        \\
VGG-11                        & 70.9  & 70.9        & 70.8        \\
VGG-16                        & 74.0  & 74.1        & 74.1        \\
VGG-19                        & 75.0  & 75.0        & 75.0        \\
SUNet-128                     & 75.6  & 76.0        & 75.4        \\
SUNet-7-128                   & 76.4  & 76.5        & 76.3        \\
DRN26                         & 75.2  & 75.3        & 75.3        \\
DRN-105                       & 79.4  & 79.5        & 79.4        \\ 
\bottomrule
\end{tabular}
\end{table}

\vspace{-5mm}
\subsection{Image Classification Networks}

Image classification networks are trained in a single phase, thus retraining simply repeats the training step schedule (with the exact same hyper-parameters and learning rate schedule as used to train the network) starting with the network initialized to its pruned trained weights.

Table~\ref{tab:classification_nets} shows the accuracy of a wide variety of networks: popular networks like ResNet~\cite{DBLP:journals/corr/HeZRS15}, VGG~\cite{Simonyan15} and Inception~\cite{DBLP:journals/corr/SzegedyLJSRAEVR14}, stacked U-Nets (SUNet)~\cite{shah2018sunets}, dilated residual networks (DRN)~\cite{DBLP:journals/corr/YuKF17}.
We also examine networks trained with weakly (WSL) or semi-weakly supervised learning (SWSL) methods~\cite{wsl,swsl} which use additional data to improve accuracy.
We prune all the convolution and fully-connected layers except for the first one ($7\times7$ convolution on 3-channel input) since they do not have a GEMM-K dimension that is an even multiple of 16.

\begin{table}[tbp]
\setlength{\tabcolsep}{5pt}
\centering
\caption{Small, parameter-efficient networks that struggle with the baseline fine-tuning approach match the target accuracy when permuting before fine-tuning (ILSVRC2012).}
\label{tab:permutations}
\renewcommand{\arraystretch}{1.10}
    \begin{tabular}{@{}lccc@{}}
    \multirow{4}{*}{\textbf{Network}} & \multicolumn{3}{c}{\textbf{Accuracy (FP16)}} \\ \cmidrule(l){2-4}
                         & \multirow{2}{*}{\textbf{Dense}}   & \multicolumn{2}{c}{\textbf{2:4 Sparse}} \\ \cmidrule(l){3-4}
                         &                  & \textbf{Default}   & \textbf{Permuted} \\
     \midrule
     MobileNet v2 & ~~~~~~71.55~~~~~~ & 69.56 & 71.56 \\
     SqueezeNet v1.0 & 58.09 & 54.08 & 58.38 \\
     SqueezeNet v1.1 & 58.21 & 56.96 & 58.23 \\
     MNASNet 1.0~~~~~~~~~~~~~ & 73.24 & 71.99 & 73.27 \\
     ShuffleNet v2 & 68.32 & 66.97 & 68.42 \\
     EfficientNet B0 & 77.25 & 75.98 & 77.29 \\
     EfficientNet-WideSE B0~~~~~~~~~~& 77.63 & 76.64 & 77.63 \\
     \bottomrule
    \end{tabular}
\end{table}

Weight and input activation tensors in convolution layers, including the first layer, and fully-connected layers are quantized to INT8.
Entropy and max calibration~\cite{haoly} are used for activations and weights, respectively.
Per-tensor scaling factors are used for activation tensors, per-channel scaling factors are used for weight tensors in convolutions, and per-row scaling factors are used for weight tensors in fully-connected layers.

The results in Table~\ref{tab:classification_nets} indicate the accuracy is maintained for both floating point and quantized networks when sparsity workflow from Section~\ref{sec:sw} is applied.
While for some networks, sparse and quantized models accuracy is slightly different than for the dense non-quantized counterparts, these differences are within bounds of run-to-run variation caused by random seeds or fine-tuning non-determinism.

Some of the lower-parameter networks (MobileNet~v2~\cite{MobileNetV2}, SqueezeNet~\cite{squeezenet}, MNASNet~\cite{mnasnet}, ShuffleNet~v2~\cite{shufflenetv2}, and EfficientNet~\cite{efficientnet}) do not fully recover accuracy when applying the basic workflow.
As Table~\ref{tab:permutations} shows, permuting the weights before pruning allows a fully recovery of accuracy for these models. 

\subsection{Image Segmentation and Detection Networks}
To study object detection and segmentation, we use networks from PyTorch Torchvision, Detectron2~\cite{wu2019detectron2}, NVIDIA Deep Learning Examples for Tensor Cores~\cite{nvidia-dl-examples}, and NVIDIA ADLR~\cite{semantic_cvpr19} repositories.

Image segmentation and object detection models are typically trained in two phases: first a backbone is trained for image classification, followed by the addition of model components (segmentation/detection heads, FPN, etc.), and then training for detection or segmentation.
Backbones are trained on the ILSVRC2012 dataset, downstream tasks are trained on COCO 2017~\cite{coco}, with some semantic segmentation networks also using Mapillary and Cityscapes datasets.

Since these detection and segmentation datasets are relatively large, we find that we can prune the weights after the second phase and repeat the training of only the second phase.
Accuracy results are summarized in Table~\ref{tab:segdet_nets}, which shows sparse results matching those of the dense counterparts.

For each network, all layers in the backbone (except the very first 3-channel convolution) and heads are pruned with \SPARSITY sparsity, 
and all layers (including the first convolution) are quantized to INT8.
Similar to classification networks, entropy calibration with per-tensor scaling factors are used for activation tensors, and max calibration with per-channel or per-row scaling factors are used for weight tensors.

\begin{table}[tbp]
\centering
\caption{Accuracy of detection (bbox Average Precision, top) and image segmentation (mask Average Precision, bottom)  networks with \SPARSITY sparsity on the COCO 2017 dataset.
In this table, RN=ResNet, FPN=Feature Pyramid Network, and RPN=Region Proposal Network.
1x and 3x are the length of fine-tuning schedules as referred to in Detectron2.
}
\label{tab:segdet_nets}
\setlength{\tabcolsep}{9pt}
\renewcommand{\arraystretch}{1.10}
\begin{tabular}{@{}llll@{}}
\multirow{3}{*}{\textbf{Network}} & \multicolumn{3}{c}{\textbf{Accuracy}}  \\  \cmidrule(l){2-4}
                          & \textbf{Dense}   & \textbf{Sparse}   & \textbf{Sparse} \vspace{-1.0mm} \\
                          & \textbf{FP16}    & \textbf{FP16}     & \textbf{INT8}    \\
\midrule
MaskRCNN-RN50                     & 37.9  & 37.9   & 37.8   \\
RetinaNet-RN50                    & 34.8  & 35.2   & 35.1  \\
SSD-RN50                          & 24.8  & 24.8   & 24.9   \\
FasterRCNN-RN50-FPN-1x            & 37.6  & 38.6   & 38.4   \\
FasterRCNN-RN50-FPN-3x            & 39.8  & 39.9   & 39.4   \\
FasterRCNN-RN101-FPN-3x           & 41.9  & 42.0   & 41.8   \\
MaskRCNN-RN50-FPN-1x              & 39.9  & 40.3   & 40.0   \\
MaskRCNN-RN50-FPN-3x              & 40.6  & 40.7   & 40.4   \\
MaskRCNN-RN101-FPN-3x             & 42.9  & 43.2   & 42.8   \\
RetinaNet-RN50-FPN-1x             & 36.4  & 37.4   & 37.2   \\
RPN-RN50-FPN-1x                   & 45.8  & 45.6   & 45.5   \\
\midrule
MaskRCNN-RN50                     & 34.5  & 34.6   & 34.5   \\
MaskRCNN-RN50-FPN-3x              & 36.9  & 37.0   & -   \\
MaskRCNN-RN101-FPN-3x~~~~~~~~~~   & 38.7  & 38.9   & -   \\
ResNeXt-101* & 81.4  & 81.7   & 81.2     \\ 
\bottomrule
\multicolumn{4}{l}{$^*$\scriptsize IoU of network pre-trained on Mapillary and fine-tuned on Cityscapes dataset} 
\end{tabular}
\end{table}

\begin{table}[h]
\setlength{\tabcolsep}{9pt}
\caption{FID scores (lower is better) of GANs with \SPARSITY sparsity.}
\centering
\label{tab:gans}
\renewcommand{\arraystretch}{1.0}
\resizebox{\columnwidth}{!}{
\begin{tabular}{@{}lllcc@{}}
\multirow{2}{*}{\textbf{Task}} & \multirow{2}{*}{\textbf{Network}} & \multirow{2}{*}{\textbf{Dataset}} & \multicolumn{2}{c}{\textbf{FID score}} \\ \cmidrule(l){4-5}
                         & & & \textbf{Dense} & \textbf{\SPARSITY Sparse} \\
\midrule
Image Synthesis & DCGAN & MNIST & 50.39 & 50.54 \\
Domain Translation & Pix2Pix & Sat $\xrightarrow{}$ Map & 17.64 & 17.89 \\
Domain Translation & Pix2Pix & Sat $\xleftarrow{}$ Map & 30.83 & 30.72 \\
Style Transfer & CycleGAN & Monet $\xrightarrow{}$ Photo & 63.15 & 63.00 \\
Style Transfer & CycleGAN & Monet $\xleftarrow{}$ Photo & 31.99 & 32.36 \\
Image-Image Translation & CycleGAN & Zebra $\xrightarrow{}$ Horse & 60.93 & 61.03 \\
Image-Image Translation & CycleGAN & Zebra $\xleftarrow{}$ Horse & 52.86 & 52.45 \\
Image-Image Translation & StarGAN & CelebA & 6.11 & 6.93 \\
Super Resolution & SRGAN & DIV2K & 14.65 & 16.60 \\
\bottomrule
\end{tabular}
} 
\end{table}

\subsection{Generative Adversarial Networks (GANs)}

As part of devising a scheme for stabilizing the fine-tuning of sparse GANs~\cite{selfsupervisedGAN}, we apply the procedure from Section~\ref{sec:sw} to generate 2:4 sparse floating-point networks to a variety of GANs and tasks.
The results for Frechet Inception Distance (FID) scores~\cite{FID} (lower is better) from this work are shown in Table~\ref{tab:gans}.

\subsection{Networks for Natural Language Processing (NLP)}

To study the behavior of our workflow on NLP tasks, we select recurrent-based translation network (GNMT~\cite{gnmt}), Transformer-based translation network (FairSeq Transformer~\cite{ott2019fairseq}), and two language modeling networks (Transformer-XL~\cite{xformerxl} and BERT~\cite{bert}). 

\subsubsection{Language Translation:}
Language translation networks are trained in a single-phase, thus the training session is repeated using the original hyper-parameters after pruning.
After fine-tuning, the networks are 
quantized to INT8 using max calibration~\cite{haoly}. Per-tensor scaling factors are used for activation tensors and per-row scaling factors are used for weight tensors in fully-connected and recurrent layers.
Table~\ref{tab:lang_xlation_nets} shows the accuracy of networks retrained for the \SPARSITY pattern matching that of the dense originals.

\begin{table}[t]
\centering
\caption{Accuracy on En-De WMT’14 of two network architectures for language translation. 
BLEU score is reported for accuracy in the table.
}
\label{tab:lang_xlation_nets}
\renewcommand{\arraystretch}{1.0}
\setlength{\tabcolsep}{8pt}
\begin{tabular}{@{}llll@{}}
\multirow{3}{*}{\textbf{Network}} & \multicolumn{3}{c}{\textbf{Accuracy}}
\\  \cmidrule(l){2-4}
                                  & \textbf{Dense} & \textbf{Sparse} & \textbf{Sparse} \vspace{-0.5mm}\\ 
                                  & \textbf{FP16}  & \textbf{FP16}   & \textbf{INT8}  \\
\midrule
GNMT                      & 24.6  & 24.9   & 24.9   \\
FairSeq Transformer       & 28.2  & 28.5   & 28.3   \\
\bottomrule
\end{tabular}
\vspace{-3mm}
\end{table}

\begin{table}[b]
\renewcommand{\arraystretch}{1.0}
\setlength{\tabcolsep}{7pt}
\centering
\caption{Accuracy of dense and sparse 12-layer Transformer-XL models in bits per character (BPC) on enwik8 and BERT\textsubscript{LARGE} F1 score on SQuAD v1.1.}
\label{tab:languagemodels}
\begin{tabular}{@{}llcccc@{}}
\multirow{3}{*}{\textbf{Network}} & \multirow{3}{*}{\textbf{Metric}} & \multicolumn{4}{c}{\textbf{Accuracy}}  \\  \cmidrule(l){3-6}
 & & \textbf{Dense} & \textbf{Sparse} & \textbf{Dense} & \textbf{Sparse} \\
 & & \textbf{FP16} & \textbf{FP16} & \textbf{INT8} & \textbf{INT8} \\
\midrule
12-layer Transformer-XL & BPC & 1.06 & 1.06 & - & - \\ 
BERT\textsubscript{LARGE} & F1 & 91.9 & 91.9 & 90.9 & 90.8 \\
\bottomrule
\end{tabular}
\end{table}

\subsubsection{Language Modeling:}
The Enwik8~\cite{enwik} dataset is used for Transformer-XL evaluation, and SQuAD v1.1~\cite{DBLP:journals/corr/RajpurkarZLL16} is used for BERT evaluation.
State-of-art language modeling networks involve a two-stage training process: 
unsupervised training (also called pre-training) on large-scale unlabeled data sets, followed by fine-tuning on much smaller data sets for 
downstream tasks, such as question-answering and entailment classification. 
For these networks, our studies show that repeating pre-training step after pruning 
and then fine-tuning the sparse network on target task 
leads to a model that matches dense network's accuracy (shown in Table~\ref{tab:languagemodels}).
We use the BERT\textsubscript{LARGE} model and training scripts as described in NVIDIA Megatron repository~\cite{megatron-repo}.
For both networks, all GEMM layers involving weight/parameter tensors inside a transformer block/layer are pruned (attention layers are not pruned since they do not involve any weights).

When quantizing BERT to INT8, all GEMM layers including batched-GEMM layers operate on INT8 operands, along with data in the residual connections.
This aggressive quantization can cause a small accuracy degradation, but sparsity matches the dense accuracy in both cases. Per-tensors scaling factors are used with max calibration for weight tensors and percentile calibration~\cite{haoly} for activations tensors in fully-connected layers. The pruned pre-trained language model is fine-tuned for both quantization operations as well as the SQuAD dataset's task-specific heads at the same time.

%% file: conclusion.tex
Sparsity in neural networks remains an active research area. Unstructured sparsity, which is the subject of many research efforts, requires very high levels of sparsity in order to achieve speedups over dense math on modern processors with matrix-math pipelines. However, very sparse networks have difficulty maintaining model accuracy. To overcome these challenges, we introduced \SPARSITY structured sparsity, hardware primitives for its acceleration, and a workflow for pruning networks. The NVIDIA Ampere GPU architecture introduces Sparse Tensor Cores, which have 2$\times$ math throughput for GEMM-like operations (convolutions and matrix multiplies) where the first argument is a tensor with \SPARSITY sparsity. The workflow was empirically shown to maintain accuracy over a wide range of tasks and neural network models, trained using standard learning rate schedules found in public code repositories.

The proposed sparsity workflow repeats a training session after pruning the weights of a trained dense networks. The benefit of this approach is that retraining does not require any hyper-parameter changes or searches. However, since this workflow doubles the training time, an interesting direction for future investigations is finding shorter fine-tuning schedules. Some preliminary experiments with a grid-search have identified a set of hyper-parameters (initial learning rate, learning rate schedule and epochs/iterations to fine-tune) for shorter fine-tuning sessions that maintained accuracy. However, these parameters were highly network- and task-dependent, and we have not yet been able to identify hyper-parameters that work universally for various tasks and networks. Thus, a universal approach to reduced fine-tuning requirements remains future work. Another interesting direction for future work is exploring the effects of sparsity on models that were trained to the limits of their capacity - there are indications that some popular models can achieve higher accuracy when trained on larger datasets or with longer training schedules. While we looked at some models trained on larger datasets (with weakly and semi-weakly supervised techniques) in Section~\ref{sec:res}, evaluating alternative training schedules is an interesting next step.

Our proposed workflow targets acceleration of inference. While this matches what would be needed to accelerate the forward pass of training, in order to also accelerate the backward pass of training, the \SPARSITY constraint must also be satisfied by the transposed weight tensors. This can be done by enforcing the constraint along both dimensions of the weight matrix: rows \textit{and} columns. The PyTorch ASP~\cite{ASP} library provides a simple greedy approach, as well as an exhaustive search, that seeks to minimize the weight magnitude lost by pruning; investigation of more efficient mask-finding algorithms is an active research area (for example,~\cite{hubara2021accelerated}) and is left as future work. To accelerate training, one would also aim to minimize the number of updates performed with a dense network. For this, one may need to train with dynamic sparsity masks, evolving them during training~\cite{learningNM} (as opposed to static masks computed once during pruning, which this paper shows to suffice for inference). Thus, investigation of dynamic mask requirements is an intriguing area for future work as well.

Finally, it is also interesting to investigate pruning of activations, as some layers, such as multi-head attention in Transformer-based networks, do not involve any weights. Our preliminary experiments suggest that it is possible to prune activations with a \SPARSITY pattern without accuracy loss; a fully general methodology is future work.